\pdfoutput=1
\documentclass[10pt,twocolumn,letterpaper]{article}

\usepackage{cvpr}
\usepackage{times}
\usepackage{epsfig}
\usepackage{graphicx}
\usepackage{amsmath}
\usepackage{amssymb}
\usepackage{subfigure}
\usepackage{float}
\usepackage{stackengine}
\usepackage{fancyhdr}
\usepackage[pagebackref=true,breaklinks=true,letterpaper=true,colorlinks,bookmarks=false]{hyperref}

\cvprfinalcopy % *** Uncomment this line for the final submission

\def\cvprPaperID{004} % *** Enter the CVPR Paper ID here

\begin{document}
\newcommand{\rulesep}{\unskip\ \vrule\ }
%%%%%%%%% TITLE
\title{Explaining the Unexplained: A CLass-Enhanced Attentive Response (CLEAR) Approach to Understanding Deep Neural Networks}

\author{Devinder Kumar\\
University of Waterloo\\
{\tt\small devinder.kumar@uwaterloo.ca}
\and
Alexander Wong\\
University of Waterloo\\
{\tt\small a28wong@uwaterloo.ca}
\and
Graham W. Taylor\\
University of Guelph \& CIFAR\\
{\tt\small gwtaylor@uoguelph.ca}
}

\maketitle
%\blfootnote{Accepted at CVPR Workshop on Explainable Computer Vision, 2017}

%%%%%%%%% ABSTRACT
\begin{abstract}
In this work, we propose \textbf{CL}ass-\textbf{E}nhanced \textbf{A}ttentive \textbf{R}esponse (CLEAR): an approach to visualize and understand the decisions made by deep neural networks (DNNs) given a specific input. CLEAR facilitates the visualization of attentive regions and levels of interest of DNNs during the decision-making process. It also enables the visualization of the most dominant classes associated with these attentive regions of interest. As such, CLEAR can mitigate some of the shortcomings of heatmap-based methods associated with decision ambiguity, and allows for better insights into the decision-making process of DNNs. Quantitative and qualitative experiments across three different datasets demonstrate the efficacy of CLEAR for gaining a better understanding of the inner workings of DNNs during the decision-making process.
\end{abstract}

%%%%%%%%% BODY TEXT
\section{Introduction}
\label{intro}

\begin{figure}[h]
\begin{center}

  \includegraphics[trim = 2cm 5cm 2.8cm 1cm ,height = 7cm,width=1.0\linewidth]{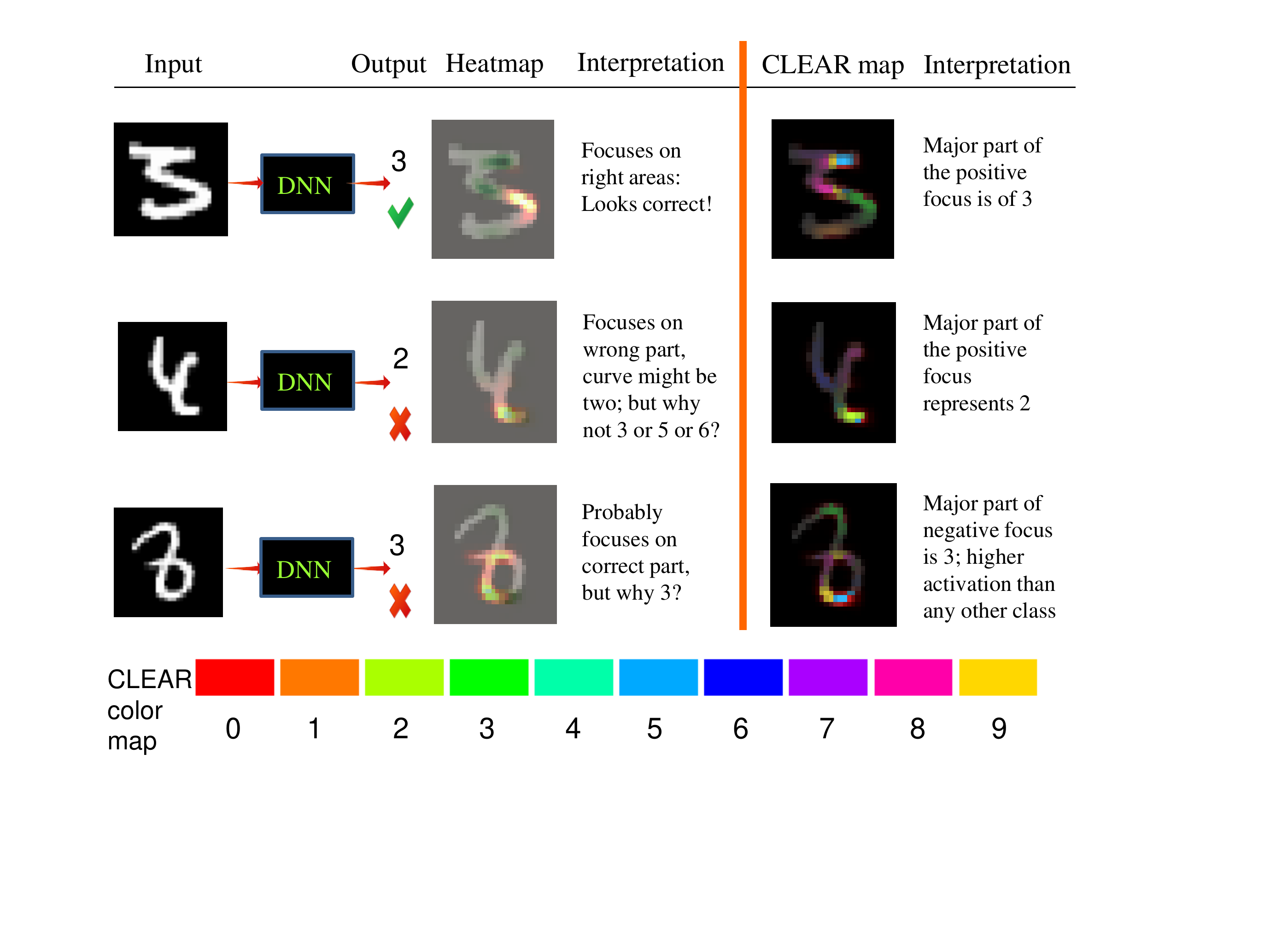}
\end{center}
   \caption{Examples of handwritten digits from MNIST are shown, along with: 1) the decision made by the DCNN, 2) heatmaps used in existing visualization methods, 3) the proposed \textbf{CL}ass-\textbf{E}nhanced \textbf{A}ttentive \textbf{R}esponse (CLEAR) maps, and 4) what can be interpreted based on the heatmaps and the proposed CLEAR maps. While the heatmaps used in existing approaches show which information in the image works for (positive focus: hot regions) or against (negative focus: green) a particular decision made, the proposed CLEAR map allows for the visualization of the attentive regions of interest, the corresponding attentive levels, as well as the dominant class for each attentive region of interest that the DCNN uses during the decision-making process. Each individual color in the CLEAR map represents the corresponding dominant attentive class at that location. Correspondence between colors and the dominant classes can be derived by the color map given at the bottom. In these examples, it is evident that the heatmaps are insufficient to fully interpret and explain the decision made by the DCNN, whereas the proposed CLEAR maps can explain the decision-making process more effectively through a multi-factor visualization approach.}
\label{fig:motive}
\end{figure}

In recent years, we have seen tremendous success in the field of artificial intelligence (AI). In particular, many of the recent advances have been related to one particular area of machine learning: deep neural networks (DNNs). DNNs have been shown to outperform previous machine learning techniques for a variety of tasks, such as fine-grained classification~\cite{zhou2016learning,cui2016fine}, self-driving cars~\cite{bojarski2016end}, captioning and answering questions about images~\cite{malinowski2015ask,antol2015vqa}, and even defeating human champions at Go~\cite{silver2016mastering}. Although DNNs have demonstrated tremendous effectiveness at a wide range of tasks, when they fail, they often fail spectacularly, producing unexplainable and incoherent results that can leave one to wonder what caused the DNN to make such decisions. This lack of transparency and interpretability of DNNs during the decision-making process is largely due to the complex nature of DNNs, where individual neural responses, unlike other interpretable decision-making processes such as decision trees, provide very little insight as to what is actually going on.

The lack of transparency in the decision-making process of DNNs is a significant bottleneck in their widespread adoption in industry, such as healthcare, defence, cyber-security, etc., where the error tolerance is very low and the ability to interpret, understand, and trust decisions is critical. As such, a way to peer inside a DNN and see why it made a decision the way it did can have tremendous potential for pushing towards explainable AI, where a human expert gains the ability to understand, interpret and verify the decisions made.

Recently, a number of researchers have been exploring the understanding and interpretation of decisions made by DNNs, in particular by Deep Convolutional Neural Networks (DCNN),  by asking the following question: \textit{based on what information in the image is the DCNN making a decision?} To tackle this question, much recent work has focused on understanding the decision-making process of networks by leveraging heatmaps that provide information about which areas of the image is used by the DCNN to make a particular decision. These approaches have produced some promising results in revealing what is important to a decision made by a DCNN. More details regarding the relevant works are provided in Section~\ref{lit_review}. A common limitation with such heatmap-based approaches to understanding the decision-making process of DCNNs is that of \textbf{decision ambiguity}, where one can gain insight into \textbf{which} regions of interest are important for making decisions, but gives no insight as to \textbf{why} such regions of interest are important. As a result, these methods leave the ``thought process'' of the DCNN largely ambiguous.

\begin{figure*}[t!]
\begin{center}
   \includegraphics[trim = 0cm 3cm 0cm 0cm ,height = 9cm,width=0.9\linewidth]{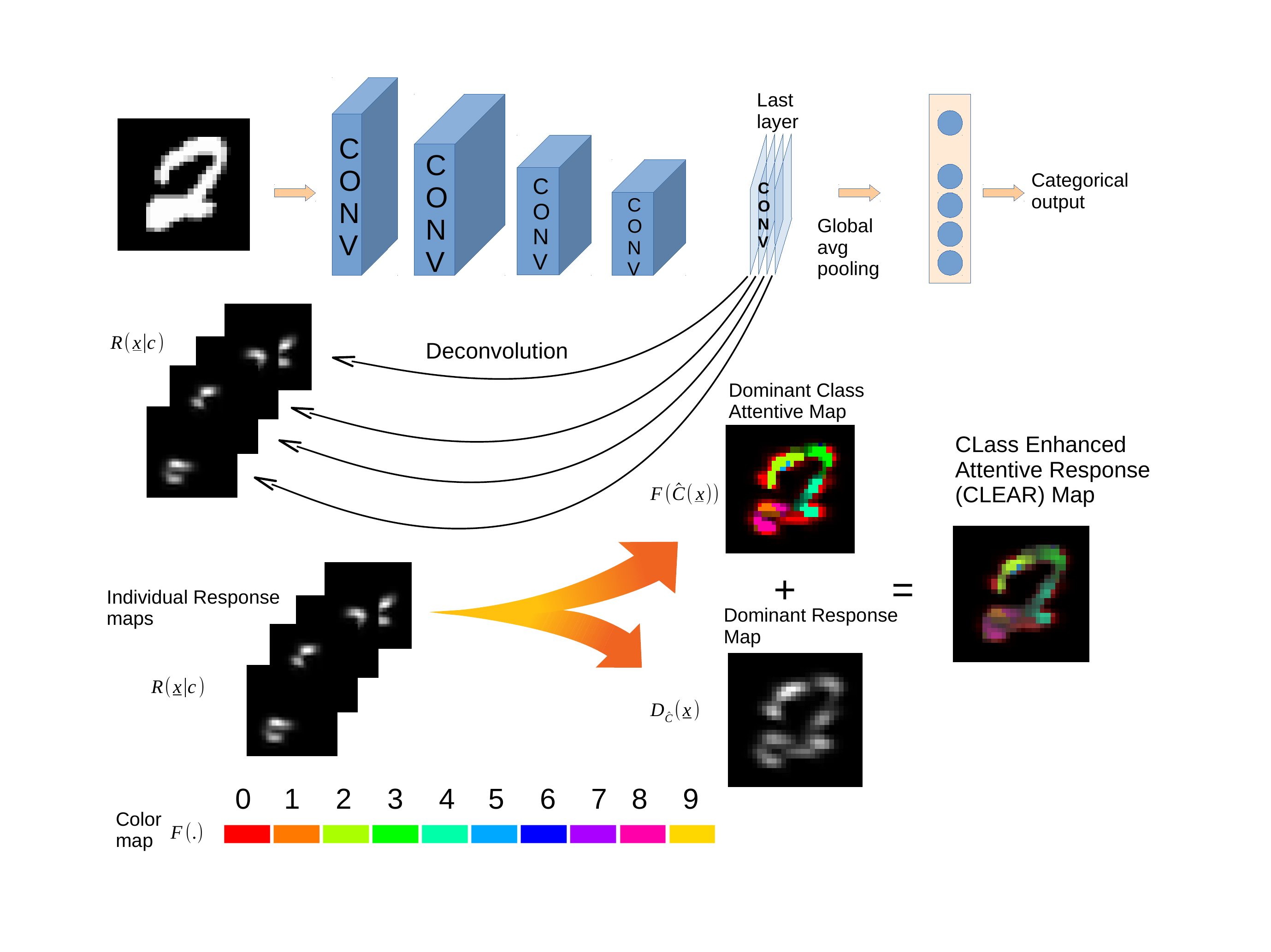}
\end{center}
   \caption{The procedure for generating \textbf{CL}ass-\textbf{E}nhanced \textbf{A}ttentive \textbf{R}esponse (CLEAR) maps. First, individual attentive response maps are computed for each class based on the last layer of the DCNN. Based on this set of attentive response maps, two different types of maps are computed: 1) a dominant attentive response map, which shows the dominant attentive level for each location in the image, and 2) a dominant class attentive map, which shows the dominant class involved in the decision-making process at each location. Finally, the dominant attentive response map and the dominant attentive class map are combined to produce the final CLEAR map for a given image.}
\label{fig:clap_explain}

\end{figure*}

In an attempt to mitigate the problem of decision ambiguity, we take a step towards ``explaining the unexplained'', with regards to the decision-making process of DCNNs, through the introduction of \textbf{CL}ass-\textbf{E}nhanced \textbf{A}ttentive \textbf{R}esponse (CLEAR) maps that go beyond what existing heatmap-based approaches~\cite{zintgraf2017visualizing,bach2015pixel,montavon2017explaining} can provide. The proposed CLEAR maps allow for the visualization of not only the attentive regions of interest and corresponding attentive levels of DCNNs during the decision-making process, but also the corresponding dominant classes associated with these attentive regions of interest. As such, compared to heatmaps, CLEAR maps are much more effective at conveying where and why certain regions of interest influence the decision-making process. An example of this is shown in Fig.~\ref{fig:motive}. We further demonstrate the effectiveness of the proposed CLEAR maps, both quantitatively and qualitatively, by conducting a number of experiments using three different publicly available datasets.
%-------------------------------------------------------------------------

\section{Related Work}
\label{lit_review}
There has been a significant body of work in recent years in the domain of visualizing and understanding DCNNs. This literature can be broadly divided into two groups: first, approaches that focus on understanding the global structure of a trained network~\cite{baehrens2010explain,goodfellow2009measuring,yosinski2015understanding}; and second, approaches that mainly focus on understanding the decision-making process of trained networks for a specific instance~\cite{springenberg2014striving,bach2015pixel,zhou2016learning,montavon2017explaining,simonyan2013deep,zeiler2014visualizing,zintgraf2017visualizing}. Our present work can be considered as belonging to the second category. Relevant work pertaining to each group is explained below:
\newline
\textbf{Global Understanding-Based Methods:} Many methods in this domain try to understand the decision-making process of the deep network by measuring its operating characteristics; for example, finding an input that maximizes the response of a particular neuron~\cite{erhan2010understanding}, measuring the network's invariance to certain kinds of data augmentation~\cite{goodfellow2009measuring}, or determining global decision structure~\cite{baehrens2010explain}. Other methods seek to find image instances from a database that maximally activate particular neurons or the posterior class probability of a given network~\cite{yosinski2015understanding}.
\newline
\textbf{Instance-Based Methods:} These methods are based on interpreting individual decisions made by a DCNN for a particular image instance. One specific instance-based method was proposed by Simonyan et al.~in~\cite{simonyan2013deep}, where using back-propagated partial derivatives of the class score with respect to pixel values were used to create class saliency maps. Zeiler \& Fergus~\cite{zeiler2014visualizing} proposed a deconvolution-based method to project the activations from feature space back to the input space (pixels) recursively. However, the method did not provide any meaning to the assignments other than that they should form a coherent set of interpretable pixels. Springenberg et al.~\cite{springenberg2014striving} provided another gradient-based visualization method, which restricts the negative gradients from flowing backwards towards the input layer, leading to sharper visualization, still without attributing any meaning to the obtained visualization. However, the study strongly showed the efficacy of networks with global average pooling for image classification and visualization. To visually discern unique features for a particular category of image, Zhou et. al.~\cite{zhou2016learning} created a \textit{class activation map} using DCNNs with global average pooling layers. This \textit{class activation map} was also used for localising objects within the image. Bach et al.~\cite{bach2015pixel} and Montavon et al.~\cite{montavon2017explaining} aimed at finding a general approach to visualize non-linear classifiers, leading to interesting heatmap generation. Recently, similar to the occlusion-based methodology for creating heatmaps in~\cite{zeiler2014visualizing}, Zintgraf et al.~\cite{zintgraf2017visualizing} proposed a method based on multivariate conditional sampling over image patches to visualize and interpret individual decisions of DCNNs as binary saliency maps to represent information that contributes for or against the decision.

In our work, instead of only obtaining feature maps, we attribute meaning to each pixel in the back-projected response in the input space using a class-based approach. Also, unlike ~\cite{bach2015pixel,montavon2017explaining} or ~\cite{zintgraf2017visualizing}, that provide heatmaps or binary heatmaps for correctly classified samples, we create CLEAR maps that are more interpretable (Fig.~\ref{fig:motive}) for both correctly or misclassified cases. Finally, compared to the per-class maps created in~\cite{zhou2016learning}, CLEAR maps show multiple class-specific contributions at once.
%-------------------------------------------------------------------------
\section{Class Enhanced Attentive Response (CLEAR)}
\label{approach}
This section explains the procedure for generating the proposed \textbf{CL}ass-\textbf{E}nhanced \textbf{A}ttentive \textbf{R}esponse (CLEAR) maps. The main goal of CLEAR maps is to convey the following information: 1) the attentive regions of interest in the image responsible for the decision made by the DCNN; 2) the attentive levels at these regions of interest so that we understand their level of influence over the decision made by the DCNN; and 3) the dominant class associated with these attentive regions of interest so that we can better understand \textbf{why} a decision was made. The procedure for generating CLEAR maps can be summarized as follows (see Fig.~\ref{fig:clap_explain}). First, individual attentive response maps are computed for each kernel associated with a class by back-projecting activations from the output layer of the DCNN. Based on this set of attentive response maps, two different types of maps are computed: 1) a dominant attentive response map, which shows the dominant attentive level for each location in the image; and 2) a dominant class attentive map, which shows the dominant class involved in the decision-making process at each location. Finally, the dominant attentive response map and the dominant attentive class map are combined visually by using color and intensity to produce the final CLEAR map for a given image.

Inspired by the effectiveness of the ALL-CNN~\cite{springenberg2014striving} on different datasets, we leveraged a similar network architecture for building the DCNN used for classification in this paper. While, for clarity, we describe the procedure for computing individual attentive response maps based on the ALL-CNN architecture, the procedure will generalize to other DCNNs provided class-specific responses can be computed in input (pixel) space. ALL-CNNs are composed primarily of convolutional, ReLU, and max-pooling layers. Towards the output of the DCNN, the last convolutional layer contains a set of kernels equal to the number of classes, and then global averaging is performed before passing these energy values to the softmax output layer which represents categories. As such, each kernel can be thought of as being associated with a particular class.

The first step of CLEAR is to compute a set of individual attentive response maps, one for each of the classes learned by the DCNN, which we will denote as $\left\{R(\underline{x}\lvert c)|1 \leq c \leq N\right\}$, where $N$ is the number of classes. This is achieved in the current realization of CLEAR by back-propagating the responses of each kernel in the last convolutional layer from feature space to the input space to form each attentive response map, thus extending upon the idea first introduced in~\cite{zeiler2010deconvolutional}. To explain the formulation for the formation of CLEAR maps, first consider a single layer of a DCNN. Let $\hat h_l$ be the deconvolved output response of the single layer $l$ with $K$ kernel weights $w$. The deconvolution output response at layer $l$ then can be then obtained by convolving each of the feature maps $z_{l}$ with kernels $w_{l}$ and summing them as:

\begin{equation}
\hat h_{l} = \sum_{k=1}^K z_{k,l} * w_{k,l} .\
\end{equation}

Here $*$ represents the convolution operation. For notational brevity, we can combine the convolution and summation operation for layer $l$ into a single convolution matrix $G_{l}$. Hence the above equation can be denoted as: $\hat h_{l} = G_{l}z_{l}$.

For multi-layered DCNNs, we can extend the above formulation by adding an additional un-pooling operation $U$ as described in~\cite{zeiler2010deconvolutional}. Thus, we can calculate the deconvolved output response from feature space to input space for any layer $l$ in a multi-layer network as:

\begin{equation}
 R_{l} = G_{1}U_{1}G_{2}U_{2} ....G_{l-1}U_{l-1}G_{l}z_{l}
\end{equation}

For CLEAR maps, we specifically calculate the output responses from individual kernels of the last layer of a network. Hence, given a network with last layer $L$ containing $K=N$ kernels, we can calculate the attentive response map; $R(\underline{x}\lvert c)$ (where $\underline{x}$ denotes the response back-projected to the input layer, and thus an array the same size as the input) for any class-specific kernel $c$ (${1 \leq c \leq N}$) in the last layer as:

\begin{equation}
 {R(\underline{x}\lvert c)} = G_{1}U_{1}G_{2}U_{2} ....G_{L-1}U_{L-1}G_{L}^c z_{L} .\
\end{equation}

Here $G_{L}^c$ represents the convolution matrix operation in which the kernel weights $w_{L}$ are all zero except that at the $c$\textsuperscript{th} location.

Given the set of individual attentive response maps, we then compute the dominant attentive class map, $\hat{C}(\underline{x})$, by finding the class at each pixel that maximizes the attentive response level, $R(\underline{x}\lvert c)$, across all classes:

\begin{equation}
 \hat{C}(\underline{x}) = \operatornamewithlimits{argmax}\limits_{c} {R(\underline{x} \lvert c)} .\
\end{equation}
Given the dominant attentive class map, $\hat{C}(\underline{x})$, we can now compute the dominant attentive response map, $D_{\hat{C}}(\underline{x})$, by selecting the attentive response level at each pixel based on the identified dominant class, which can be expressed as follows:
\begin{equation}
 D_{\hat{C}}(\underline{x}) = R(\underline{x}\lvert \hat{C}) .\
\end{equation}
To form the final CLEAR map, we map the dominant class attentive map and the dominant attentive response map in the HSV color space as follows, then transform back into the RGB color space:
\begin{equation}
\begin{split}
   H & = F(\hat{C}(\underline{x})) ,\ \\
   S & = 1 ,\ \\
   V & = D_{\hat{C}}(\underline{x}) .\
\end{split}
\end{equation}

Here $F(.)$ is the color map dictionary that assigns an individual color to each dominant attentive class, $c$. Fig.~\ref{fig:clap_explain} shows an example of the CLEAR map overlayed on the image.

%-----------------------------------------------------------------------
\section{Experiments}
\label{exp}
In this section, we illustrate the efficacy of CLEAR maps for understanding and interpreting the decision-making of DCNNs. We conducted qualitative and quantitative experiments on three different datasets: the commonly used benchmarks MNIST and Street View House Numbers (SVHN), and the Stanford Dog dataset~\cite{KhoslaYaoJayadevaprakashFeiFei_FGVC2011}. In the following section, we explain the experimental setup.

\subsection{Setup}
To conduct experiments on three different datasets, we trained three different DCNN architectures with all convolutional layers\footnote{Network architectures are shown in Appendix.}. For training on MNIST and SVHN, we set our network architecture similar to~\cite{springenberg2014striving}, as it has shown to perform very effectively for a variety of datasets. To train these networks, we used the default train and test split. We achieved an accuracy of $99.26\%$ and $92.6\%$ for the MNIST and SVHN datasets, respectively. For training on the Stanford dog dataset, we used a 16 layer VGG net~\cite{simonyan2014very}, pre-trained on ImageNet. We modified the VGG net slightly by removing the two last fully-connected layers and augmenting with two convolutional layers at the end. We fine-tuned the last two layers using the Stanford dog dataset. As with MNIST and SVHN, the default train and test split was applied to this dataset, but instead, we only took 10 different classes for training. We made this decision, as it would be an arduous task to interpret from all 120 classes. For this fine-grained classification task, we achieved an accuracy of 58.74\% for 10 classes, whereas the state-of-the art for this dataset with 120 classes is 68\%.

In all three networks, as the last layer (convolutional layer) was linearly connected to the softmax activation function, each kernel can be considered to represent one separate class. It is important to note that the aim was to understand and interpret the decision of a trained network; hence we did not strive to achieve the best architecture and state-of-the art results for each dataset. Using the previously mentioned setup, we conducted the following experiments.

\begin{figure}[htb!]
\centering
\begin{center}
    \stackunder[3pt]{\label{fig:0}\includegraphics[height=0.4cm,width=0.04\textwidth]{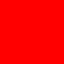}}{0}
    \stackunder[3pt]{\label{fig:0}\includegraphics[height=0.4cm,width=0.04\textwidth]{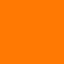}}{1}
    \stackunder[3pt]{\label{fig:0}\includegraphics[height=0.4cm,width=0.04\textwidth]{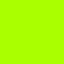}}{2}
    \stackunder[3pt]{\label{fig:0}\includegraphics[height=0.4cm,width=0.04\textwidth]{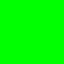}}{3}
   \stackunder[3pt]{\label{fig:0}\includegraphics[height=0.4cm,width=0.04\textwidth]{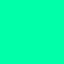}}{4}
   \stackunder[3pt]{\label{fig:0}\includegraphics[height=0.4cm,width=0.04\textwidth]{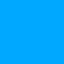}}{5}
    \stackunder[3pt]{\label{fig:0}\includegraphics[height=0.4cm,width=0.04\textwidth]{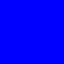}}{6}
   \stackunder[3pt]{\label{fig:0}\includegraphics[height=0.4cm,width=0.04\textwidth]{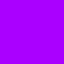}}{7}
    \stackunder[3pt]{\label{fig:0}\includegraphics[height=0.4cm,width=0.04\textwidth]{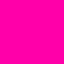}}{8}
    \stackunder[3pt]{\label{fig:0}\includegraphics[height=0.4cm,width=0.04\textwidth]{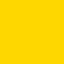}}{9}
  \end{center}
  \begin{center}
       \hspace{-0.6cm} Correctly classified \hspace{0.8 cm}\vline \hspace{1.2 cm} Misclassified \\
       I \hspace{1cm} II \hspace{1cm} III  \hspace{- 0.2cm} \hspace{1cm} I \hspace{1cm} II \hspace{1cm} III
    \end{center}
  \includegraphics[trim = 0cm 0cm 0cm 0cm,height = 1.3 cm ,width=0.07\textwidth]{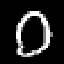}
  \includegraphics[trim = 0cm 0cm 0cm 0cm,height = 1.3 cm ,width=0.07\textwidth]{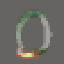}
  \includegraphics[trim = 0cm 0cm 0cm 0cm,height = 1.3 cm ,width=0.07\textwidth]{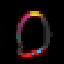}
  \includegraphics[trim = 0cm 0cm 0cm 0cm,height = 1.3 cm ,width=0.07\textwidth]{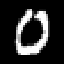}
  \includegraphics[trim = 0cm 0cm 0cm 0cm,height = 1.3 cm ,width=0.07\textwidth]{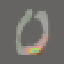}
  \includegraphics[trim = 0cm 0cm 0cm 0cm,height = 1.3 cm ,width=0.07\textwidth]{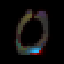}
  \includegraphics[trim = 0cm 0cm 0cm 0cm,height = 1.3 cm ,width=0.07\textwidth]{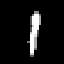}
  \includegraphics[trim = 0cm 0cm 0cm 0cm,height = 1.3 cm ,width=0.07\textwidth]{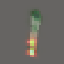}
  \includegraphics[trim = 0cm 0cm 0cm 0cm,height = 1.3 cm ,width=0.07\textwidth]{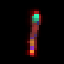}
\includegraphics[trim = 0cm 0cm 0cm 0cm,height = 1.3 cm ,width=0.07\textwidth]{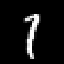}
  \includegraphics[trim = 0cm 0cm 0cm 0cm,height = 1.3 cm ,width=0.07\textwidth]{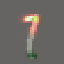}
  \includegraphics[trim = 0cm 0cm 0cm 0cm,height = 1.3 cm ,width=0.07\textwidth]{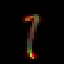}

  \includegraphics[trim = 0cm 0cm 0cm 0cm,height = 1.3 cm ,width=0.07\textwidth]{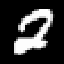}
  \includegraphics[trim = 0cm 0cm 0cm 0cm,height = 1.3 cm ,width=0.07\textwidth]{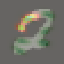}
  \includegraphics[trim = 0cm 0cm 0cm 0cm,height = 1.3 cm ,width=0.07\textwidth]{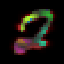}
    \includegraphics[trim = 0cm 0cm 0cm 0cm,height = 1.3 cm ,width=0.07\textwidth]{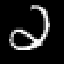}
  \includegraphics[trim = 0cm 0cm 0cm 0cm,height = 1.3 cm ,width=0.07\textwidth]{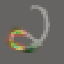}
  \includegraphics[trim = 0cm 0cm 0cm 0cm,height = 1.3 cm ,width=0.07\textwidth]{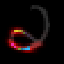}
   \includegraphics[trim = 0cm 0cm 0cm 0cm,height = 1.3 cm ,width=0.07\textwidth]{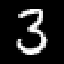}
  \includegraphics[trim = 0cm 0cm 0cm 0cm,height = 1.3 cm ,width=0.07\textwidth]{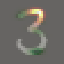}
  \includegraphics[trim = 0cm 0cm 0cm 0cm,height = 1.3 cm ,width=0.07\textwidth]{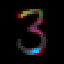}
    \includegraphics[trim = 0cm 0cm 0cm 0cm,height = 1.3 cm ,width=0.07\textwidth]{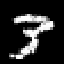}
  \includegraphics[trim = 0cm 0cm 0cm 0cm,height = 1.3 cm ,width=0.07\textwidth]{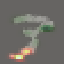}
  \includegraphics[trim = 0cm 0cm 0cm 0cm,height = 1.3 cm ,width=0.07\textwidth]{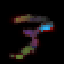}

  \includegraphics[trim = 0cm 0cm 0cm 0cm,height = 1.3 cm ,width=0.07\textwidth]{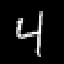}
  \includegraphics[trim = 0cm 0cm 0cm 0cm,height = 1.3 cm ,width=0.07\textwidth]{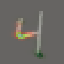}
  \includegraphics[trim = 0cm 0cm 0cm 0cm,height = 1.3 cm ,width=0.07\textwidth]{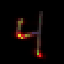}
\includegraphics[trim = 0cm 0cm 0cm 0cm,height = 1.3 cm ,width=0.07\textwidth]{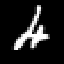}
  \includegraphics[trim = 0cm 0cm 0cm 0cm,height = 1.3 cm ,width=0.07\textwidth]{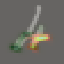}
  \includegraphics[trim = 0cm 0cm 0cm 0cm,height = 1.3 cm ,width=0.07\textwidth]{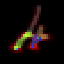}
  \includegraphics[trim = 0cm 0cm 0cm 0cm,height = 1.3 cm ,width=0.07\textwidth]{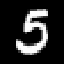}
  \includegraphics[trim = 0cm 0cm 0cm 0cm,height = 1.3 cm ,width=0.07\textwidth]{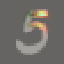}
  \includegraphics[trim = 0cm 0cm 0cm 0cm,height = 1.3 cm ,width=0.07\textwidth]{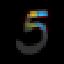}
   \includegraphics[trim = 0cm 0cm 0cm 0cm,height = 1.3 cm ,width=0.07\textwidth]{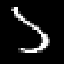}
  \includegraphics[trim = 0cm 0cm 0cm 0cm,height = 1.3 cm ,width=0.07\textwidth]{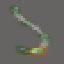}
  \includegraphics[trim = 0cm 0cm 0cm 0cm,height = 1.3 cm ,width=0.07\textwidth]{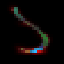}
  \includegraphics[trim = 0cm 0cm 0cm 0cm,height = 1.3 cm ,width=0.07\textwidth]{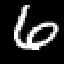}
  \includegraphics[trim = 0cm 0cm 0cm 0cm,height = 1.3 cm ,width=0.07\textwidth]{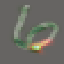}
  \includegraphics[trim = 0cm 0cm 0cm 0cm,height = 1.3 cm ,width=0.07\textwidth]{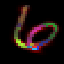}
     \includegraphics[trim = 0cm 0cm 0cm 0cm,height = 1.3 cm ,width=0.07\textwidth]{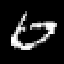}
  \includegraphics[trim = 0cm 0cm 0cm 0cm,height = 1.3 cm ,width=0.07\textwidth]{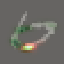}
  \includegraphics[trim = 0cm 0cm 0cm 0cm,height = 1.3 cm ,width=0.07\textwidth]{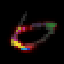}
  \includegraphics[trim = 0cm 0cm 0cm 0cm,height = 1.3 cm ,width=0.07\textwidth]{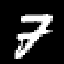}
  \includegraphics[trim = 0cm 0cm 0cm 0cm,height = 1.3 cm ,width=0.07\textwidth]{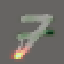}
  \includegraphics[trim = 0cm 0cm 0cm 0cm,height = 1.3 cm ,width=0.07\textwidth]{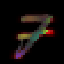}
  \includegraphics[trim = 0cm 0cm 0cm 0cm,height = 1.3 cm ,width=0.07\textwidth]{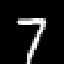}
  \includegraphics[trim = 0cm 0cm 0cm 0cm,height = 1.3 cm ,width=0.07\textwidth]{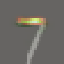}
  \includegraphics[trim = 0cm 0cm 0cm 0cm,height = 1.3 cm ,width=0.07\textwidth]{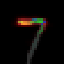}

  \includegraphics[trim = 0cm 0cm 0cm 0cm,height = 1.3 cm ,width=0.07\textwidth]{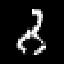}
  \includegraphics[trim = 0cm 0cm 0cm 0cm,height = 1.3 cm ,width=0.07\textwidth]{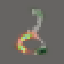}
  \includegraphics[trim = 0cm 0cm 0cm 0cm,height = 1.3 cm ,width=0.07\textwidth]{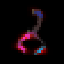}
   \includegraphics[trim = 0cm 0cm 0cm 0cm,height = 1.3 cm ,width=0.07\textwidth]{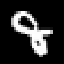}
  \includegraphics[trim = 0cm 0cm 0cm 0cm,height = 1.3 cm ,width=0.07\textwidth]{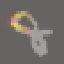}
  \includegraphics[trim = 0cm 0cm 0cm 0cm,height = 1.3 cm ,width=0.07\textwidth]{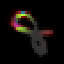}

  \includegraphics[trim = 0cm 0cm 0cm 0cm,height = 1.3 cm ,width=0.07\textwidth]{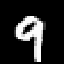}
  \includegraphics[trim = 0cm 0cm 0cm 0cm,height = 1.3 cm ,width=0.07\textwidth]{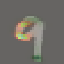}
  \includegraphics[trim = 0cm 0cm 0cm 0cm,height = 1.3 cm ,width=0.07\textwidth]{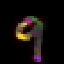}
  \includegraphics[trim = 0cm 0cm 0cm 0cm,height = 1.3 cm ,width=0.07\textwidth]{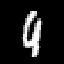}
  \includegraphics[trim = 0cm 0cm 0cm 0cm,height = 1.3 cm ,width=0.07\textwidth]{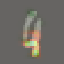}
  \includegraphics[trim = 0cm 0cm 0cm 0cm,height = 1.3 cm ,width=0.07\textwidth]{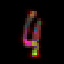}
  \caption{Example images from the MNIST dataset. Each row represents two sets of examples for digit 0-9: correctly classified example (left) and misclassified example (right). Each example set consists of the (I) original image, (II) heatmap results (where hot regions are focus of positive kernel and green represents dominant pixel results for the rest of kernels in the last layer) and (III) CLEAR maps. The color map on top shows the associations of different colors with their respective classes in the CLEAR map.}
  \label{fig:mnist}
\end{figure}

\subsection{Qualitative Experiments: Understanding The Decision Making Process }
In this set of experiments, we first create binary heatmaps and the proposed CLEAR maps for individual images in the three different datasets. The binary heatmaps represent which information in the image was used for or against the true class versus other image classes during classification. The binary heatmaps were formed by overlaying the output response from the kernel representing the true class as ``hot'' regions and response of the rest of the kernels in the last layer, represented by green regions. The response for rest of the kernels is formed by performing max operation across the individual output responses. Thus, in the binary heatmaps, the hot regions and green regions represent the information for and against the actual class respectively, that was used for decision-making by the network. The binary heatmaps are constructed similarly to ~\cite{zintgraf2017visualizing} and ~\cite{montavon2017explaining}. The CLEAR map formation is explained in Section~\ref{approach} and Fig.~\ref{fig:clap_explain}.

\begin{figure*}[ht]
\centering
\begin{center}
     \stackunder[3pt]{\label{fig:0}\includegraphics[height=0.4cm,width=0.08\textwidth]{fig/color_map/color_map_img_index_0.png}}{0}
     \stackunder[3pt]{\label{fig:0}\includegraphics[height=0.4cm,width=0.08\textwidth]{fig/color_map/color_map_img_index_1.png}}{1}
      \stackunder[3pt]{\label{fig:0}\includegraphics[height=0.4cm,width=0.08\textwidth]{fig/color_map/color_map_img_index_2.png}}{2}
      \stackunder[3pt]{\label{fig:0}\includegraphics[height=0.4cm,width=0.08\textwidth]{fig/color_map/color_map_img_index_3.png}}{3}
     \stackunder[3pt]{\label{fig:0}\includegraphics[height=0.4cm,width=0.08\textwidth]{fig/color_map/color_map_img_index_4.png}}{4}
      \stackunder[3pt]{\label{fig:0}\includegraphics[height=0.4cm,width=0.08\textwidth]{fig/color_map/color_map_img_index_5.png}}{5}
      \stackunder[3pt]{\label{fig:0}\includegraphics[height=0.4cm,width=0.08\textwidth]{fig/color_map/color_map_img_index_6.png}}{6}
     \stackunder[3pt]{\label{fig:0}\includegraphics[height=0.4cm,width=0.08\textwidth]{fig/color_map/color_map_img_index_7.png}}{7}
      \stackunder[3pt]{\label{fig:0}\includegraphics[height=0.4cm,width=0.08\textwidth]{fig/color_map/color_map_img_index_8.png}}{8}
      \stackunder[3pt]{\label{fig:0}\includegraphics[height=0.4cm,width=0.08\textwidth]{fig/color_map/color_map_img_index_9.png}}{9}
  \end{center}
  \begin{center}
      Correctly classified \hspace{02.1cm} \vline \hspace{03cm} Misclassified\\
	I \hspace{1.5cm} II \hspace{1.5cm} III \hspace{1.5cm} IV \hspace{0.2cm} \hspace{1.5cm} I \hspace{1.5cm} II \hspace{1.5cm} III \hspace{1.5cm} IV
    \end{center}
  \includegraphics[trim = 0cm 0cm 0cm 0cm,height = 1 cm ,width=0.11\textwidth]{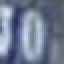}
  \includegraphics[trim = 0cm 0cm 0cm 0cm,height = 1 cm ,width=0.11\textwidth]{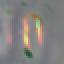}
  \includegraphics[trim = 0cm 0cm 0cm 0cm,height = 1 cm ,width=0.11\textwidth]{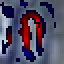}
  \includegraphics[trim = 0cm 0cm 0cm 0cm,height = 1 cm ,width=0.11\textwidth]{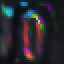}
  \includegraphics[trim = 0cm 0cm 0cm 0cm,height = 1 cm ,width=0.11\textwidth]{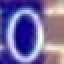}
  \includegraphics[trim = 0cm 0cm 0cm 0cm,height = 1 cm ,width=0.11\textwidth]{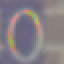}
  \includegraphics[trim = 0cm 0cm 0cm 0cm,height = 1 cm ,width=0.11\textwidth]{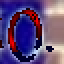}
  \includegraphics[trim = 0cm 0cm 0cm 0cm,height = 1 cm ,width=0.11\textwidth]{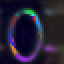}

  \includegraphics[trim = 0cm 0cm 0cm 0cm,height = 1 cm ,width=0.11\textwidth]{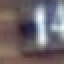}
  \includegraphics[trim = 0cm 0cm 0cm 0cm,height = 1 cm ,width=0.11\textwidth]{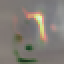}
  \includegraphics[trim = 0cm 0cm 0cm 0cm,height = 1 cm ,width=0.11\textwidth]{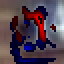}
  \includegraphics[trim = 0cm 0cm 0cm 0cm,height = 1 cm ,width=0.11\textwidth]{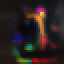}
  \includegraphics[trim = 0cm 0cm 0cm 0cm,height = 1 cm ,width=0.11\textwidth]{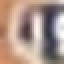}
  \includegraphics[trim = 0cm 0cm 0cm 0cm,height = 1 cm ,width=0.11\textwidth]{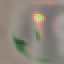}
  \includegraphics[trim = 0cm 0cm 0cm 0cm,height = 1 cm ,width=0.11\textwidth]{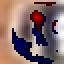}
  \includegraphics[trim = 0cm 0cm 0cm 0cm,height = 1 cm ,width=0.11\textwidth]{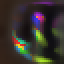}

  \includegraphics[trim = 0cm 0cm 0cm 0cm,height = 1 cm ,width=0.11\textwidth]{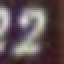}
  \includegraphics[trim = 0cm 0cm 0cm 0cm,height = 1 cm ,width=0.11\textwidth]{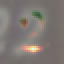}
  \includegraphics[trim = 0cm 0cm 0cm 0cm,height = 1 cm ,width=0.11\textwidth]{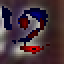}
  \includegraphics[trim = 0cm 0cm 0cm 0cm,height = 1 cm ,width=0.11\textwidth]{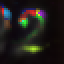}
  \includegraphics[trim = 0cm 0cm 0cm 0cm,height = 1 cm ,width=0.11\textwidth]{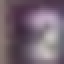}
  \includegraphics[trim = 0cm 0cm 0cm 0cm,height = 1 cm ,width=0.11\textwidth]{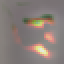}
  \includegraphics[trim = 0cm 0cm 0cm 0cm,height = 1 cm ,width=0.11\textwidth]{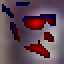}
  \includegraphics[trim = 0cm 0cm 0cm 0cm,height = 1 cm ,width=0.11\textwidth]{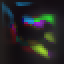}

  \includegraphics[trim = 0cm 0cm 0cm 0cm,height = 1 cm ,width=0.11\textwidth]{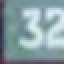}
  \includegraphics[trim = 0cm 0cm 0cm 0cm,height = 1 cm ,width=0.11\textwidth]{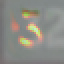}
  \includegraphics[trim = 0cm 0cm 0cm 0cm,height = 1 cm ,width=0.11\textwidth]{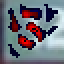}
  \includegraphics[trim = 0cm 0cm 0cm 0cm,height = 1 cm ,width=0.11\textwidth]{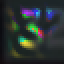}
  \includegraphics[trim = 0cm 0cm 0cm 0cm,height = 1 cm ,width=0.11\textwidth]{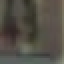}
  \includegraphics[trim = 0cm 0cm 0cm 0cm,height = 1 cm ,width=0.11\textwidth]{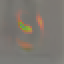}
  \includegraphics[trim = 0cm 0cm 0cm 0cm,height = 1 cm ,width=0.11\textwidth]{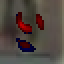}
  \includegraphics[trim = 0cm 0cm 0cm 0cm,height = 1 cm ,width=0.11\textwidth]{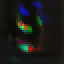}

  \includegraphics[trim = 0cm 0cm 0cm 0cm,height = 1 cm ,width=0.11\textwidth]{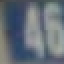}
  \includegraphics[trim = 0cm 0cm 0cm 0cm,height = 1 cm ,width=0.11\textwidth]{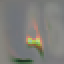}
  \includegraphics[trim = 0cm 0cm 0cm 0cm,height = 1 cm ,width=0.11\textwidth]{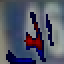}
  \includegraphics[trim = 0cm 0cm 0cm 0cm,height = 1 cm ,width=0.11\textwidth]{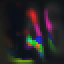}
  \includegraphics[trim = 0cm 0cm 0cm 0cm,height = 1 cm ,width=0.11\textwidth]{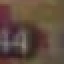}
  \includegraphics[trim = 0cm 0cm 0cm 0cm,height = 1 cm ,width=0.11\textwidth]{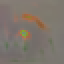}
  \includegraphics[trim = 0cm 0cm 0cm 0cm,height = 1 cm ,width=0.11\textwidth]{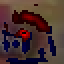}
  \includegraphics[trim = 0cm 0cm 0cm 0cm,height = 1 cm ,width=0.11\textwidth]{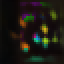}

  \includegraphics[trim = 0cm 0cm 0cm 0cm,height = 1 cm ,width=0.11\textwidth]{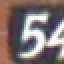}
  \includegraphics[trim = 0cm 0cm 0cm 0cm,height = 1 cm ,width=0.11\textwidth]{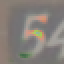}
  \includegraphics[trim = 0cm 0cm 0cm 0cm,height = 1 cm ,width=0.11\textwidth]{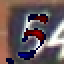}
  \includegraphics[trim = 0cm 0cm 0cm 0cm,height = 1 cm ,width=0.11\textwidth]{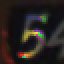}
  \includegraphics[trim = 0cm 0cm 0cm 0cm,height = 1 cm ,width=0.11\textwidth]{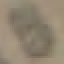}
  \includegraphics[trim = 0cm 0cm 0cm 0cm,height = 1 cm ,width=0.11\textwidth]{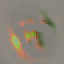}
  \includegraphics[trim = 0cm 0cm 0cm 0cm,height = 1 cm ,width=0.11\textwidth]{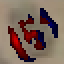}
  \includegraphics[trim = 0cm 0cm 0cm 0cm,height = 1 cm ,width=0.11\textwidth]{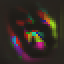}

  \includegraphics[trim = 0cm 0cm 0cm 0cm,height = 1 cm ,width=0.11\textwidth]{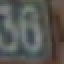}
  \includegraphics[trim = 0cm 0cm 0cm 0cm,height = 1 cm ,width=0.11\textwidth]{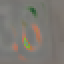}
  \includegraphics[trim = 0cm 0cm 0cm 0cm,height = 1 cm ,width=0.11\textwidth]{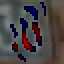}
  \includegraphics[trim = 0cm 0cm 0cm 0cm,height = 1 cm ,width=0.11\textwidth]{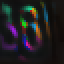}
  \includegraphics[trim = 0cm 0cm 0cm 0cm,height = 1 cm ,width=0.11\textwidth]{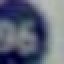}
  \includegraphics[trim = 0cm 0cm 0cm 0cm,height = 1 cm ,width=0.11\textwidth]{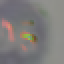}
  \includegraphics[trim = 0cm 0cm 0cm 0cm,height = 1 cm ,width=0.11\textwidth]{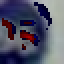}
  \includegraphics[trim = 0cm 0cm 0cm 0cm,height = 1 cm ,width=0.11\textwidth]{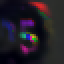}

  \includegraphics[trim = 0cm 0cm 0cm 0cm,height = 1 cm ,width=0.11\textwidth]{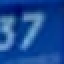}
  \includegraphics[trim = 0cm 0cm 0cm 0cm,height = 1 cm ,width=0.11\textwidth]{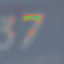}
  \includegraphics[trim = 0cm 0cm 0cm 0cm,height = 1 cm ,width=0.11\textwidth]{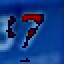}
  \includegraphics[trim = 0cm 0cm 0cm 0cm,height = 1 cm ,width=0.11\textwidth]{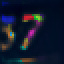}
  \includegraphics[trim = 0cm 0cm 0cm 0cm,height = 1 cm ,width=0.11\textwidth]{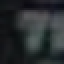}
  \includegraphics[trim = 0cm 0cm 0cm 0cm,height = 1 cm ,width=0.11\textwidth]{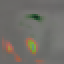}
  \includegraphics[trim = 0cm 0cm 0cm 0cm,height = 1 cm ,width=0.11\textwidth]{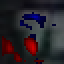}
  \includegraphics[trim = 0cm 0cm 0cm 0cm,height = 1 cm ,width=0.11\textwidth]{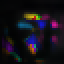}

  \includegraphics[trim = 0cm 0cm 0cm 0cm,height = 1 cm ,width=0.11\textwidth]{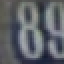}
  \includegraphics[trim = 0cm 0cm 0cm 0cm,height = 1 cm ,width=0.11\textwidth]{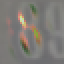}
  \includegraphics[trim = 0cm 0cm 0cm 0cm,height = 1 cm ,width=0.11\textwidth]{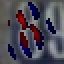}
  \includegraphics[trim = 0cm 0cm 0cm 0cm,height = 1 cm ,width=0.11\textwidth]{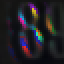}
  \includegraphics[trim = 0cm 0cm 0cm 0cm,height = 1 cm ,width=0.11\textwidth]{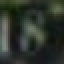}
  \includegraphics[trim = 0cm 0cm 0cm 0cm,height = 1 cm ,width=0.11\textwidth]{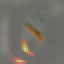}
  \includegraphics[trim = 0cm 0cm 0cm 0cm,height = 1 cm ,width=0.11\textwidth]{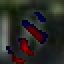}
  \includegraphics[trim = 0cm 0cm 0cm 0cm,height = 1 cm ,width=0.11\textwidth]{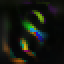}
  \includegraphics[trim = 0cm 0cm 0cm 0cm,height = 1 cm ,width=0.11\textwidth]{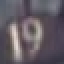}
  \includegraphics[trim = 0cm 0cm 0cm 0cm,height = 1 cm ,width=0.11\textwidth]{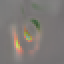}
  \includegraphics[trim = 0cm 0cm 0cm 0cm,height = 1 cm ,width=0.11\textwidth]{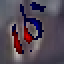}
  \includegraphics[trim = 0cm 0cm 0cm 0cm,height = 1 cm ,width=0.11\textwidth]{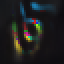}
  \includegraphics[trim = 0cm 0cm 0cm 0cm,height = 1 cm ,width=0.11\textwidth]{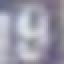}
  \includegraphics[trim = 0cm 0cm 0cm 0cm,height = 1 cm ,width=0.11\textwidth]{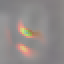}
  \includegraphics[trim = 0cm 0cm 0cm 0cm,height = 1 cm ,width=0.11\textwidth]{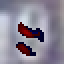}
  \includegraphics[trim = 0cm 0cm 0cm 0cm,height = 1 cm ,width=0.11\textwidth]{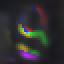}
  \caption{Correctly classified (left) and misclassified (right) images from the SVHN dataset. Each row represents two sets of examples for digit 0-9. Each example set consists of the (I) original image, (II) heatmap results (where hot regions are focus of positive kernels, and green regions for the rest of kernels), (III) binary map (red represents information for and blue represents information against the given image class) and (IV) CLEAR map respectively. The color map at the top shows the associations of different colors with their respective classes in the CLEAR map.}
  \label{fig:svhn}
\end{figure*}

\begin{figure}[ht]
\centering

\begin{center}
      \stackunder[3pt]{\label{fig:0}\includegraphics[height=0.2cm,width=0.09\textwidth]{fig/color_map/color_map_img_index_0.png}}{Chihuahua}
    \stackunder[3pt]{\label{fig:0}\includegraphics[height=0.2cm,width=0.09\textwidth]{fig/color_map/color_map_img_index_1.png}}{Jap. Spaniel}
    \stackunder[3pt]{\label{fig:0}\includegraphics[height=0.2cm,width=0.09\textwidth]{fig/color_map/color_map_img_index_2.png}}{Maltese}
    \stackunder[3pt]{\label{fig:0}\includegraphics[height=0.2cm,width=0.09\textwidth]{fig/color_map/color_map_img_index_3.png}}{Pekinese}
     \stackunder[3pt]{\label{fig:0}\includegraphics[height=0.2cm,width=0.09\textwidth]{fig/color_map/color_map_img_index_4.png}}{Shih-Tzu}
     \stackunder[3pt]{\label{fig:0}\includegraphics[height=0.2cm,width=0.09\textwidth]{fig/color_map/color_map_img_index_5.png}}{B. Spaniel}
     \stackunder[3pt]{\label{fig:0}\includegraphics[height=0.2cm,width=0.09\textwidth]{fig/color_map/color_map_img_index_6.png}}{Papillon}
     \stackunder[3pt]{\label{fig:0}\includegraphics[height=0.2cm,width=0.09\textwidth]{fig/color_map/color_map_img_index_7.png}}{Toy Terrier}
     \stackunder[3pt]{\label{fig:0}\includegraphics[height=0.2cm,width=0.09\textwidth]{fig/color_map/color_map_img_index_8.png}}{R. Ridgeback}
     \stackunder[3pt]{\label{fig:0}\includegraphics[height=0.2cm,width=0.09\textwidth]{fig/color_map/color_map_img_index_9.png}}{Afghan Hound}
  \end{center}
        Shih-Tzu \hspace{0.5cm} Maltese \hspace{0.6cm} Chihuahua \hspace{0.4cm} Toy-Terrier

  \includegraphics[trim = 0cm 0cm 0cm 0cm,height = 6 cm ,width=0.45\textwidth]{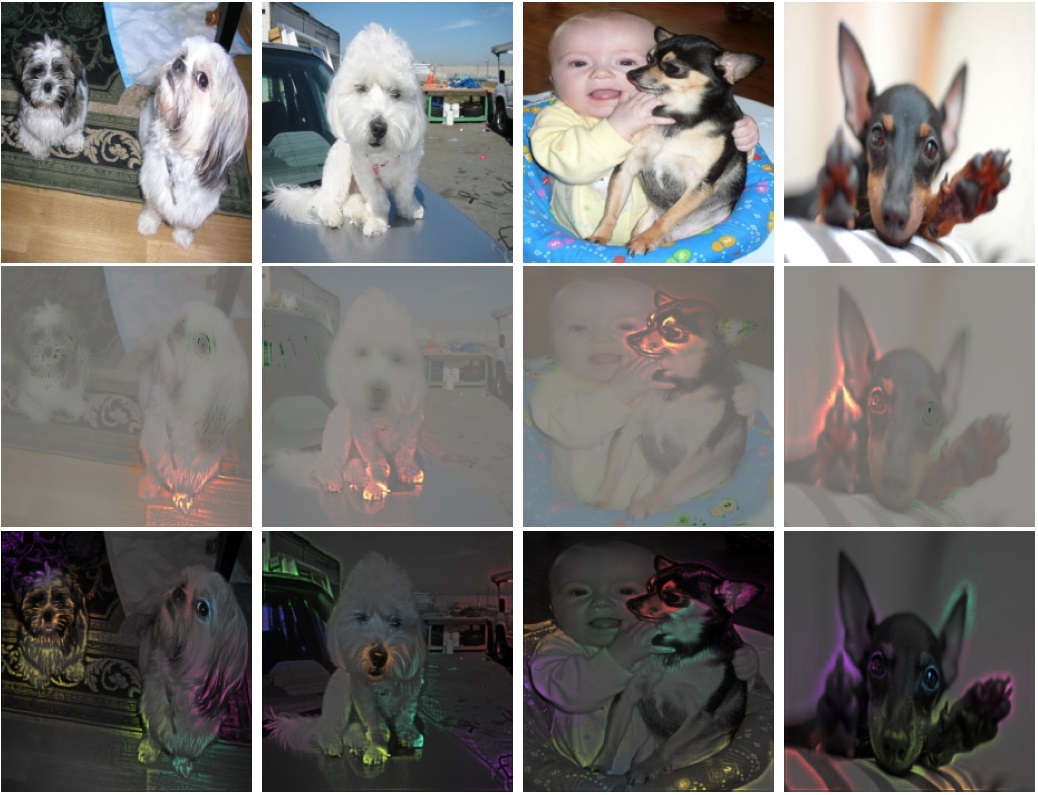}

  \caption{Multi-class maps for four different classes (Shih-Tzu, Maltese, Chihuahua and Toy-Terrier). In the last row, each color in the CLEAR maps represents a distinct class (breed) which can be identified by the color map at the top of the figure. }
  \label{fig:multi_db_eg}
\end{figure}

\begin{figure*}[h]
\centering
  \includegraphics[trim = 0cm 0cm 0cm 0cm,height = 1.5 cm ,width=0.15\textwidth]{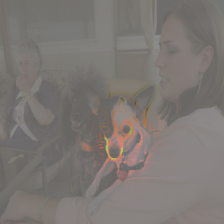}
  \includegraphics[trim = 0cm 0cm 0cm 0cm,height = 1.5 cm ,width=0.15\textwidth]{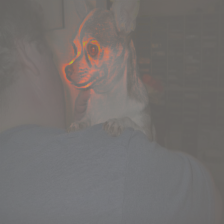}
  \includegraphics[trim = 0cm 0cm 0cm 0cm,height = 1.5 cm ,width=0.15\textwidth]{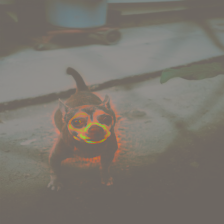}
  \rulesep\rulesep
  \includegraphics[trim = 0cm 0cm 0cm 0cm,height = 1.5 cm ,width=0.15\textwidth]{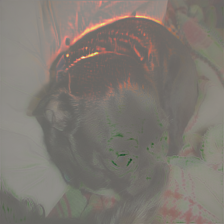}
  \includegraphics[trim = 0cm 0cm 0cm 0cm,height = 1.5 cm ,width=0.15\textwidth]{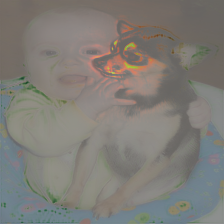}

  \includegraphics[trim = 0cm 0cm 0cm 0cm,height = 1.5 cm ,width=0.15\textwidth]{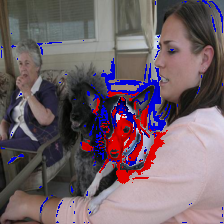}
  \includegraphics[trim = 0cm 0cm 0cm 0cm,height = 1.5 cm ,width=0.15\textwidth]{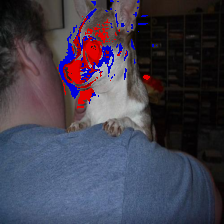}
  \includegraphics[trim = 0cm 0cm 0cm 0cm,height = 1.5 cm ,width=0.15\textwidth]{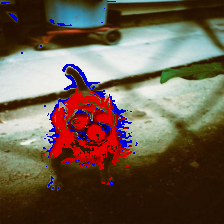}
  \rulesep\rulesep
  \includegraphics[trim = 0cm 0cm 0cm 0cm,height = 1.5 cm ,width=0.15\textwidth]{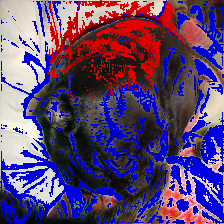}
  \includegraphics[trim = 0cm 0cm 0cm 0cm,height = 1.5 cm ,width=0.15\textwidth]{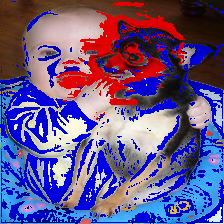}

  \includegraphics[trim = 0cm 0cm 0cm 0cm,height = 1.5 cm ,width=0.15\textwidth]{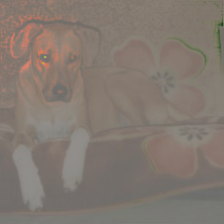}
  \includegraphics[trim = 0cm 0cm 0cm 0cm,height = 1.5 cm ,width=0.15\textwidth]{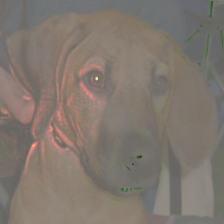}
  \includegraphics[trim = 0cm 0cm 0cm 0cm,height = 1.5 cm ,width=0.15\textwidth]{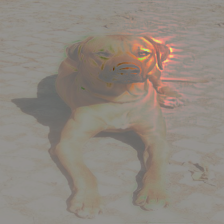}
  \rulesep\rulesep
  \includegraphics[trim = 0cm 0cm 0cm 0cm,height = 1.5 cm ,width=0.15\textwidth]{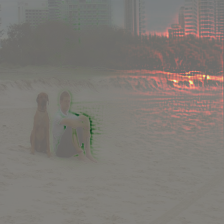}
  \includegraphics[trim = 0cm 0cm 0cm 0cm,height = 1.5 cm ,width=0.15\textwidth]{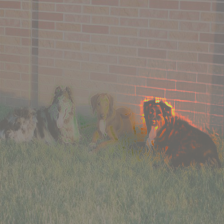}

  \includegraphics[trim = 0cm 0cm 0cm 0cm,height = 1.5 cm ,width=0.15\textwidth]{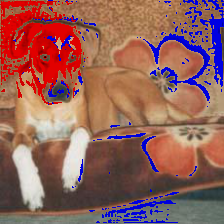}
  \includegraphics[trim = 0cm 0cm 0cm 0cm,height = 1.5 cm ,width=0.15\textwidth]{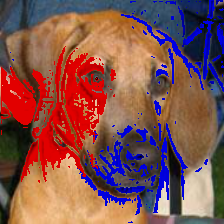}
  \includegraphics[trim = 0cm 0cm 0cm 0cm,height = 1.5 cm ,width=0.15\textwidth]{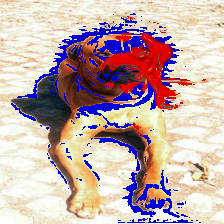}
  \rulesep\rulesep
  \includegraphics[trim = 0cm 0cm 0cm 0cm,height = 1.5 cm ,width=0.15\textwidth]{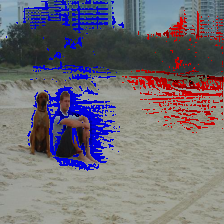}
  \includegraphics[trim = 0cm 0cm 0cm 0cm,height = 1.5 cm ,width=0.15\textwidth]{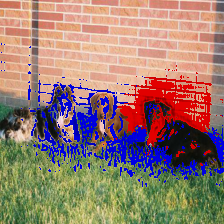}

  \caption{Three correctly classified examples (left), and two misclassified (right), for two different breeds of dogs. It can be observed that for the Chihuahua breed (top), face structure is used as a strong identifier. But we can also observe that for one of the examples (last image), even when the network focuses on correct features, it fails to classify correctly and thinks it is another breed, Shih-Tzu (for explanation, please see Fig~\ref{fig:multi_fail_db_eg}). For the Rhodesian Ridgeback breed (bottom), the face, along with ears, are a strong identifier, but it fails to identify the breed while looking at the same features again in the rightmost image.}
  \label{fig:full_db_pos_neg_eg}
\end{figure*}

\begin{figure}[h]

\begin{center}
     \stackunder[3pt]{\label{fig:0}\includegraphics[height=0.2cm,width=0.09\textwidth]{fig/color_map/color_map_img_index_0.png}}{Chihuahua}
     \stackunder[3pt]{\label{fig:0}\includegraphics[height=0.2cm,width=0.09\textwidth]{fig/color_map/color_map_img_index_1.png}}{Jap. Spaniel}
      \stackunder[3pt]{\label{fig:0}\includegraphics[height=0.2cm,width=0.09\textwidth]{fig/color_map/color_map_img_index_2.png}}{Maltese}
      \stackunder[3pt]{\label{fig:0}\includegraphics[height=0.2cm,width=0.09\textwidth]{fig/color_map/color_map_img_index_3.png}}{Pekinese}
      \stackunder[3pt]{\label{fig:0}\includegraphics[height=0.2cm,width=0.09\textwidth]{fig/color_map/color_map_img_index_4.png}}{Shih-Tzu}
     \stackunder[3pt]{\label{fig:0}\includegraphics[height=0.2cm,width=0.09\textwidth]{fig/color_map/color_map_img_index_5.png}}{B. Spaniel}
     \stackunder[3pt]{\label{fig:0}\includegraphics[height=0.2cm,width=0.09\textwidth]{fig/color_map/color_map_img_index_6.png}}{Papillon}
      \stackunder[3pt]{\label{fig:0}\includegraphics[height=0.2cm,width=0.09\textwidth]{fig/color_map/color_map_img_index_7.png}}{Toy Terrier}
     \stackunder[3pt]{\label{fig:0}\includegraphics[height=0.2cm,width=0.09\textwidth]{fig/color_map/color_map_img_index_8.png}}{R. Ridgeback}
      \stackunder[3pt]{\label{fig:0}\includegraphics[height=0.2cm,width=0.09\textwidth]{fig/color_map/color_map_img_index_9.png}}{Afghan Hound}
  \end{center}
\centering
  \includegraphics[trim = 0cm 0cm 0cm 0cm,height = 1.8 cm ,width=0.13\textwidth]{fig/dog/full/fw14.png}
  \includegraphics[trim = 0cm 0cm 0cm 0cm,height = 1.8 cm ,width=0.13\textwidth]{fig/dog/full/fwb14.png}
  \includegraphics[trim = 0cm 0cm 0cm 0cm,height = 1.8 cm ,width=0.13\textwidth]{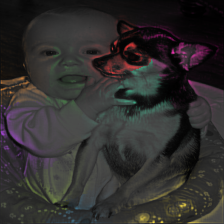}
  \includegraphics[trim = 0cm 0cm 0cm 0cm,height = 1.8 cm ,width=0.13\textwidth]{fig/dog/full/fw25.png}
  \includegraphics[trim = 0cm 0cm 0cm 0cm,height = 1.8 cm ,width=0.13\textwidth]{fig/dog/full/fwb24.png}
  \includegraphics[trim = 0cm 0cm 0cm 0cm,height = 1.8 cm ,width=0.13\textwidth]{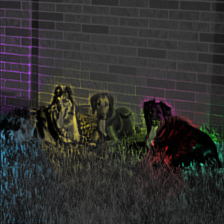}
  \caption{Heatmap, binary map, and CLEAR map for two misclassified images are shown. From the heatmap and binary map, it can be seen that the network correctly focuses on the strong identifying features of the Chihuahua and Ridgeback class, as shown in Fig~\ref{fig:full_db_pos_neg_eg}, but still fails to classify the image correctly. The CLEAR maps in the third column show that activations for the correct Chihuahua class (red) is smaller than the activation for Shih-Tzu (light blueish), hence the network thinks that it is a Shih-Tzu instead of a Chihuahua. Similarly for the Ridgeback class, the activations of Afghan Hound are higher than the true class's activation.}
  \label{fig:multi_fail_db_eg}
\end{figure}

For the SVHN and Stanford datasets, we also create an additional binary map. This map replaces the varying values in the binary heatmaps with a constant value. In the binary map, red and blue regions represent the information used for and against the class, respectively. We create these maps for visual clarity, as sometimes it is harder to visualize the green regions in the binary heatmaps.
\newline
\textbf{MNIST}: Some of the randomly chosen results for the MNIST dataset are shown in Fig.~\ref{fig:mnist}. This figure shows examples of correctly classified and misclassified examples by the network. From these results, observations that can be made are: 1) Looking at the example sets for digit 0, although positive support is contributed by the same bottom curved features in both examples, only in one case is the image correctly identified as zero. Looking at the CLEAR maps, we can see the dominant activations for the correctly classified example corresponds to class 0, whereas for the misclassified case they correspond to class 5. 2) Similarly, for digit 7 and 8 it is difficult to interpret the decision output of the DCNN, but looking at the CLEAR maps make them more interpretable.
\newline
\textbf{SVHN}: Presented similarly to the MNIST dataset, results obtained for the SVHN dataset are shown in Fig.~\ref{fig:svhn}. Some interesting observations are as follows: 1) For the misclassifed 0 digit, the heatmap overwhelmingly focuses on the correct curves; but the network still misclassifies it. This is counterintuitive to human interpretation. But when observing the CLEAR maps, we see that almost all the strong activations are for classes other than 0. 2) For the digit 9, it is difficult to interpret the binary heatmaps, as the positive kernel focuses on the digit 1, but it still correctly classifies the digit as 9 with high confidence. Observing the CLEAR maps, we see that most of the dominant activation in the focus areas belong to digit 9, including the ones for digit 1.
\newline
\textbf{Stanford Dog dataset}: Results for Stanford Dogs are shown in Fig.~\ref{fig:multi_db_eg}--\ref{fig:multi_fail_db_eg}.
The key observations are the following: 1) Binary heatmaps can be used to find strong identifying features for different classes, as shown in Fig.~\ref{fig:full_db_pos_neg_eg}. 2) In the same figure, the rightmost misclassifed cases present an interesting observation. We can observe that the network mis-classifies the Chihuahua breed as Shiz-Tzu and the Ridgeback breed as Afghan Hound. This happens even when the \textit{positive} kernels associated with their respective true class focus on the strong discriminating features, as identified by correctly classified images on the left. In Fig.~\ref{fig:multi_fail_db_eg}, the CLEAR maps show that for Chihuahuas, the strong activations are for the Shih-Tzu, whereas for the Ridgeback, the activations are stronger for Afghan Hounds.

Based on these results and observations, it is evident that binary maps are not enough for interpreting and explaining the individual decision outputs of a network. There is a strong motivation for class-based maps, such as CLEAR maps, that are more effective for understanding and interpreting the classification decisions made by a DCNN.

%-----------------

%-----------------------------------------------------------------------

%------------------------------------------------------

\subsection{Quantitative Experiments}
To re-validate our observations for the MNIST and SVHN datasets, we conducted two different quantitative experiments. In the first experiment, we removed all parts of the image, except for regions responsible for the activations of the kernel associated with the class of the image (\textit{positive kernel}). We call these regions \textit{strong features} associated with the class. For the MNIST dataset, we replace the digit with the background and for the SVHN dataset, we replace the region with a gray patch.
\begin{table}[ht]
\centering
\caption{Evaluation to re-validate the effectiveness and contribution of identified strong features on accuracy.}
	\begin{tabular}{p{4cm}p{1.5cm}p{1.5cm}}
	\hline\noalign{\smallskip}
	\hline\noalign{\smallskip}
	 \textbf{Accuracy(\%)} & \textbf{MNIST}  &\textbf{SVHN} \\
	\hline

	Full image & 99.26 & 92.60 \\
	{with only strong features} 	& 79.89 & 69.12   \\
	{without strong features} 	& 43.45 & 54.46    \\
  \hline\noalign{\smallskip}
  \hline\noalign{\smallskip}
	\end{tabular}

	\label{tab:class_result}
\end{table}
In the second experiment, we do the opposite: we remove the regions responsible for the kernel associated with the true class of the input image and keep the rest of the image. Results are shown in Table~\ref{tab:class_result}, and demonstrate that the identified strong features are vital for correctly classifying a particular class. For the case where the network is still able to classify without the strong features, albeit with half of the accuracy in comparison to the above case, an argument can be made that for these cases, the network focuses again on similar or redundant features. An example is digit 3, where there are redundant strong curve features.

\section{Discussion}
 This section discusses some general points associated with the CLEAR maps approach: 1) It is interesting to note that in Fig.~\ref{fig:clap_explain}, there is sparsity in the individual response maps from the last layer kernels. We observed the same pattern for all datasets considered. Evidence for classes tends to come from very specific localized regions. 2) For the datasets with a large number of classes, like the Stanford Dog dataset, we selected and created CLEAR maps with only 10 classes. We didn't strive to show the CLEAR maps for all 120 classes, as doing so would make it extremely difficult to interpret the decision outputs. For such cases, perhaps showing the top 10 most activated class or several different maps with $N$ classes would be a better approach. 3) In the current realization of our approach, we use deconvolution responses with only fully convolutional networks. We would like to point out that even though end-to-end learning in this case is only possible with Fully Convolutional Nets (FCN), our approach can be extended to be used with different network architectures with the use different response methods, such as Layer-wise Relevance Propagation (LRP)~\cite{bach2015pixel}, Deep Taylor decomposition~\cite{montavon2017explaining}, or prediction differential analysis~\cite{zintgraf2017visualizing}.

\section{Conclusion}
\label{conclusion}
In this work, a novel approach to better understanding and visualizing the decision-making process of DNNs was introduced in the form of \textbf{CL}ass-\textbf{E}nhanced \textbf{A}ttentive \textbf{R}esponse (CLEAR) maps. CLEAR maps are designed to enable the visualization of not only the areas of interest that predominantly influence the decision-making process, but also the degree of influence as well as the dominant class of influence in these areas. This multi-faceted look at the decision-making process allows for a better understanding of not only where but why certain decisions are made by DCNNs compared to existing heatmap-based approaches. Experiments using three different publicly available datasets were performed and show the efficacy of CLEAR maps both quantitatively and qualitatively.  Furthermore, we demonstrated that strong areas of interest identified with CLEAR maps play a pivotal role in the correct classification of the class. Future work will explore extending CLEAR to facilitate for scenarios characterized by a large number of classes (i.e.~greater than 10), as well as exploring CLEAR with different network architectures.
%------------------------------------------------------
\section*{Acknowledgment}
This research has been supported by Canada Research Chairs programs, Natural Sciences Engineering Research Council of Canada (NSERC), and Canada Foundation for Innovation (CFI).

%------------------------------------------------------
{\small
\bibliographystyle{ieee}
\bibliography{egbib}
}
%-----------------------------------------------------------
\section*{Appendix}

This section presents architectures of three different Deep Convolutional Neural Networks (DCNN) used in our paper for experiments on three different datasets as listed below:.

\subsection{MNIST}
\begin{table}[h]
	\centering
	\caption{Architecture of our DCNN used for MNIST Classification.}
%	\begin{tabular}{p{2.5cm}|p{2.5cm}|p{2.5cm}}
	\begin{tabular}{c c }
	%\begin{tabular}{|p{1.7cm}|p{4.5cm}|p{3cm}|p{3.5cm}|p{1.7cm}|}
		\hline

			 Conv Layer 							& 	(3x3, 32x) \\
			 Conv Layer 							& 	(3x3, 32x) \\
			 Conv Layer 							& 	(3x3, 32x) \\
			 Conv Layer 							& 	(3x3, 32x) \\
			 MaxPool Layer 				& 	(2x2, 2x2 stride) \\
			 Conv Layer 							& 	(3x3, 64x) \\
			 Conv Layer 							& 	(1x1, 10x) \\
			 Global average pooling							& (10) \\
			 Softmax								& 	(10) \\
		\hline
	\end{tabular}
	\label{deep_arch}
\end{table}

\subsection{SVHN}
\begin{table}[h]
	\centering
	\caption{Architecture of our DCNN used for SVHN Classification.}
%	\begin{tabular}{p{2.5cm}|p{2.5cm}|p{2.5cm}}
	\begin{tabular}{c c }
	%\begin{tabular}{|p{1.7cm}|p{4.5cm}|p{3cm}|p{3.5cm}|p{1.7cm}|}
		\hline

			 Conv Layer 							& 	(3x3, 32x) \\
			 Conv Layer 							& 	(3x3, 32x) \\
			 Conv Layer 							& 	(3x3, 32x) \\
			 MaxPool Layer 				& 	(2x2, 2x2 stride) \\
			 Conv Layer 							& 	(3x3, 64x) \\
			 Conv Layer 							& 	(3x3, 64x) \\
			 Conv Layer 							& 	(3x3, 64x) \\
			 MaxPool Layer 				& 	(2x2, 2x2 stride) \\
			 Conv Layer 							& 	(3x3, 128x) \\
			 Conv Layer 							& 	(1x1, 128x) \\
			 Conv Layer 							& 	(1x1, 10x) \\
			 Global average pooling							& (10) \\
			 Softmax								& 	(10) \\
		\hline
	\end{tabular}
	\label{deep_arch}
\end{table}
\vspace{10cm}

\subsection{Stanford Dog} 
\begin{table}[h]
	\centering
	\caption{Architecture of our DCNN used for Stanford Dog dataset Classification.}
%	\begin{tabular}{p{2.5cm}|p{2.5cm}|p{2.5cm}}

	\begin{tabular}{c c }
	%\begin{tabular}{|p{1.7cm}|p{4.5cm}|p{3cm}|p{3.5cm}|p{1.7cm}|}
	
		\hline

			 Conv Layer 							& 	(3x3, 64x) \\
			 Conv Layer 							& 	(3x3, 64x) \\
			 MaxPool Layer 				& 	(2x2, 2x2 stride) \\
			 Conv Layer 							& 	(3x3, 128x) \\
			 Conv Layer 							& 	(3x3, 128x) \\
			 MaxPool Layer 				& 	(2x2, 2x2 stride) \\
			 Conv Layer 							& 	(3x3, 256x) \\
			 Conv Layer 							& 	(3x3, 256x) \\
			 Conv Layer 							& 	(3x3, 256x) \\
			 MaxPool Layer 				& 	(2x2, 2x2 stride) \\
			 Conv Layer 							& 	(3x3, 512x) \\
			 Conv Layer 							& 	(3x3, 512x) \\
			 Conv Layer 							& 	(3x3, 512x) \\
			 MaxPool Layer 				& 	(2x2, 2x2 stride) \\ 
			 Conv Layer 							& 	(3x3, 512x) \\
			 Conv Layer 							& 	(3x3, 512x) \\
			 Conv Layer 							& 	(3x3, 512x) \\
			 MaxPool Layer 				& 	(2x2, 2x2 stride) \\ 
			 Conv Layer 							& 	(3x3, 512x) \\
			 Conv Layer 							& 	(1x1, 512x) \\
			 Conv Layer 							& 	(1x1, 10x) \\
			 Global average pooling							& (10) \\
			 Softmax								& 	(10) \\
		\hline
	\end{tabular}
	\label{deep_arch}
\end{table}

\end{document}